\documentclass[sigconf,nonacm]{acmart}

\AtBeginDocument{%
  \providecommand\BibTeX{{%
    \normalfont B\kern-0.5em{\scshape i\kern-0.25em b}\kern-0.8em\TeX}}}

\acmConference[Woodstock '18]{Woodstock '18: ACM Symposium on Neural Gaze Detection}{June 03--05, 2018}{Woodstock, NY}
\acmBooktitle{Woodstock '18: ACM Symposium on Neural Gaze Detection, June 03--05, 2018, Woodstock, NY}
\acmPrice{15.00}
\acmISBN{978-1-4503-9999-9/18/06}

\usepackage{graphicx}
\usepackage{caption}
\usepackage{subcaption}
\usepackage{amsfonts}
\usepackage{amsmath}
\usepackage[ruled,vlined]{algorithm2e}
\usepackage{bm}
\usepackage{epstopdf}
\usepackage{soul}

\usepackage{xcolor}

\definecolor{event_cyan}{HTML}{009090}
\definecolor{light_red}{HTML}{ff8080}
\definecolor{dark_red}{HTML}{d00000}
\definecolor{light_blue}{HTML}{87ceff}
\definecolor{dark_blue}{HTML}{0000b0}

\usepackage{bm}

\newcommand*{\Eqref}[1]{Eq.~\eqref{#1}}
\def\*#1{\mathbf{#1}}

\usepackage{bbold}
\usepackage{amsmath}
\usepackage{graphicx}
\usepackage[colorinlistoftodos]{todonotes}
\usepackage{dsfont}
\graphicspath{{}}

\begin{document}

\sloppy
\title{Recurrent Adversarial Service Times}

\author{C\'esar Ojeda}
\affiliation{%
  \institution{Fraunhofer Center for Machine Learning and Fraunhofer IAIS, 53757 Sankt Augustin, Germany}
}
\email{cesar.ali.ojeda.marin@iais.fraunhofer.de}

\author{Kostadin Cvejosky}
\affiliation{%
  \institution{Fraunhofer Center for Machine Learning and Fraunhofer IAIS, 53757 Sankt Augustin, Germany}
}
\email{Kostadin.cvejoski@iais.fraunhofer.de}

\author{Rams\'es J. S\'anchez}
\affiliation{%
  \institution{B-IT, University of Bonn, \\ Bonn, Germany}
}
\email{sanchez@bit.uni-bonn.de}

\author{Jannis Schuecker}
\affiliation{%
  \institution{Fraunhofer Center for Machine Learning and Fraunhofer IAIS, 53757 Sankt Augustin, Germany}
}
\email{jannis.schuecker@iais.fraunhofer.de}

\author{Bogdan Georgiev}
\affiliation{%
  \institution{Fraunhofer Center for Machine Learning and Fraunhofer IAIS, 53757 Sankt Augustin, Germany}
}
\email{bogdan.georgiev@iais.fraunhofer.de}

\author{Christian Bauckhage}
\affiliation{%
  \institution{Fraunhofer Center for Machine Learning and Fraunhofer IAIS, 53757 Sankt Augustin, Germany}
}
\email{christian.bauckhage@iais.fraunhofer.de}

\renewcommand{\shortauthors}{Ojeda, et al.}

\begin{abstract}
Service system dynamics occur at the interplay between customer behaviour and a service provider's response. This kind of dynamics can effectively be modeled within the framework of queuing theory where customers' arrivals are described by point process models. However, these approaches are limited by parametric assumptions as to, for example, inter-event time distributions. 
In this paper, we address these limitations and propose a novel, deep neural network solution to the queuing problem. Our solution combines a recurrent neural network that models the arrival process with a recurrent generative adversarial network which models the service time distribution. We evaluate our methodology on various empirical datasets ranging from internet services (Blockchain, GitHub, Stackoverflow) to mobility service systems (New York taxi cab).
\end{abstract}





\maketitle

\section{Introduction}

The ultimate success of any service provider rests on their ability to quickly and efficiently satisfy their customers: a mobility system is only successful if its users arrive on time; a block-chain is reliable provided that low latency of its transaction times is ensured; Internet services can only retain users if they provide quick and fast response. To operate systems like these, one needs to understand not only \emph{when} customers will require a service but also \emph{how} the system is able to react and respond to demands. Moreover, one should be able to dynamically adapt the system to external events. Here, examples include sudden disruptions of the mobility system due to a car crash or weather conditions or financial crises and breaking news  affecting block-chain transactions. 

Recent research has focused primarily on the customer side of a service system. For instance, research on dynamical recommender systems aims at understanding the change in users' preferences over time. These changes, together with trends as to the popularity of items, determine suitable recommendations \cite{wu2017recurrent}, \cite{jing2017neural}, \cite{zhou2018jump}. Within this line of research the customer dynamics are analyzed using point process theory and are modeled using parametric forms such as Poisson- or Hawkes processes. However, these parametric forms constrain the model's expressibility to capture the users' dynamic behavior. This drawback has lately been tackled by means of flexible non parametric models such as recurrent neural networks and Gaussian processes \cite{mei2017neural}\cite{du2016recurrent}. 

 Another line of research focuses on the service side of the system, in articular on service times. Corresponding approaches typically resort to queuing theory: a customer expresses a demand and the system decides when to serve this demand. Within queuing theory, the customer dynamics is modeled with a point process and, yet again, parametric forms are assumed. Results are usually limited to moments of the service time distribution \cite{asmussen2008applied} or are, in some cases, based on Bayesian inference \cite{sutton2011bayesian}. Yet, the latter is often neither flexible nor scalable enough to handle millions of customers in modern service systems.
 
 The work presented here aims at combining recurrent neural networks for modeling the customer arrival process with flexible service time distribution models. Our contributions are as follows:
\begin{itemize}
    \item[] \textbf{First deep solutions to service times for queuing systems}: to the best of our knowledge, this is the first approach exploiting the representation learning capabilities of deep neural networks for point processes to infer service time distributions. We provide two models: first, distributions parametrized by multilayered  perceptrons, and, second, generative adversarial neural networks. The adversarial models successfully capture multi-modal and long tail service time distributions.
    \item[] \textbf{General solutions}: our methodologies deliver holistic solutions for general families of arrival and service processes, superior to classical theoretical models which are constrained to some specific aspects of either the arrival or the service process.
    \item[] \textbf{Dynamic services}: we introduce solutions which characterize independent service time dynamics thus allowing for exogenous events to be characterized implicitly.
    \item[] \textbf{Bitcoin mempool}: to the best of our knowledge, we provide the first deep and non parametric solution for the prediction of unconfirmed transactions and block creations in the Bitcoin network.
    \item[] \textbf{Predicting and sampling from point process}: additionally, we provide a new general framework for prediction of- and sampling from recurrent point process models.
\end{itemize}

Our presentation proceeds as follows: In Section \ref{sec: background} we present the theoretical basis of our models, which are then introduced in Section \ref{sec: DeepServiceTimes}. Section \ref{sec: experiments} presents extensive empirical evaluations of our approach and Section \ref{sec: discussion} finally concludes this paper.
% \cite{Gallager}\cite{Kleinrock}.

\section{Background}
\label{sec: background}

In this section we introduce the basic concept of queuings and the standard notation of the theory. We also introduce the Recurrent Point Process model (RPP) \cite{du2016recurrent}, with which we model the customer arrival process and which will serve as a starting point for our service time models. Finally, we provide a new framework for prediction and sampling from the RPP model.

\subsection{Queues Notation}

The theory of queuings deals with the study of customer's service times in a system. One would like to know how much time a customer is likely to wait for a service, or how much time this service is expected to last. It corresponds to a central technique in the area of operations research, as it is of fundamental interest to efficiently allocate time and resources in a given service system. Historically, the field emerged from the studies related to telephone exchange and call arrivals.

In order to define a queuing system one must specify the nature of the client arrivals as well as the system server dynamics. One system might allow for a finite amount of clients which can be served at a time, or the service times might dynamically change with every incoming client. In the initial framework, one models the arrivals as a point process, traditionally specifying the inter-arrival time (i.e. the difference between two consecutive arrivals) of this process. In the following we denote the arrival times as a sequence $a_i\in\mathbb{R^+}$. After arrival, the client waits until the service dynamics chooses to start the corresponding service. This waiting time is denoted as $w_i\in\mathbb{R^+}$. After the service is completed the client leaves the systems at a departure time, denoted by $d_i\in\mathbb{R^+}$. These departure times also define a point process, to which we refer in the following as \textit{departure process}. The service time $s_i\in\mathbb{R^+}$ is defined as the amount of time the $i$th client spends being served, i.e. $s_i=d_i-a_i$.

The standard notation used to specify the characteristics of the different queuing systems consists of characters separated by slashes: $\cdot/\cdot/\cdot$. The first character describes the arrival process, namely the inter arrival distribution. Typical examples are  $M$ for memoryless (Poisson), $D$ for deterministic times and $G$ for general distributions. The second character specifies the service process, i.e. the service time distribution, and the third one specifies the number of servers available in the system. For example, a $M/M/1$ queuing corresponds to a queuing system with Poisson arrivals and exponential distributed service times for a single server. 

In this work we provide a general solution to the $G/G/\infty$ queuing problem.

\subsection{Recurrent Point Process}
\label{ssec:RMPP}
% The RPP model provides a rich representation of the customer arrival dynamics, modeling conditional intensity functions with non parametric state transitions via recurrent neural networks.
In what follows we write the likelihood of a Poisson process induced by an intensity function $\lambda^{*}(t)$ and defined through a recurrent neural network, following the procedure stated in \cite{NeuralHawkes}, \cite{RecurrentTemporal}. Let us consider a point process with a compact support $\mathcal{S} \subset \mathbb{R}$. Formally, the likelihood is written down as an inhomogeneous Poisson process between arrivals conditioned on the history of arrivals $\mathcal{H}_j \equiv  \{a_1,...,a_j\}$\footnote{a.k.a. filtration} \cite{VereJones}. For one-dimensional processes the conditional likelihood function reads
\begin{equation}
f^*(t) = \lambda^*(t) \exp\left\{ \int^{t}_{a_j}\lambda^*\left(t'\right)dt' \right\}\, ,
\label{eq:point_likelihood}
\end{equation}
where $\lambda^*$ is (locally) integrable function.
The functional dependence of the intensity function is given by a recurrent neural network (RNN) with hidden state $\*h_j$, where an exponential function guarantees that the intensity is \textit{non-negative}
\begin{equation}
\lambda^*(t) = \exp{\left\{\*v^t\cdot\*h_j + w^t\left(t-a_j\right) + b^t\right\}} \label{eq:intensity}\, .
\end{equation}
Here the vector $\*v^t$ and the scalars $w^t$ and $b^t$ are trainable variables. We remark that although recurrent networks \cite{GravesRecurrent} are defined over \textit{sequences}, the point process likelihood \Eqref{eq:point_likelihood} requires evaluation of the function over the whole support $\mathcal{S}$. In both \cite{NeuralHawkes} and \cite{RecurrentTemporal} this problem is bypassed by defining decaying continuous values between two arrivals $a_j$ and $a_{j+1}$, either for \textit{memory cells} in LSTMs or hidden layers. 

The update equation for the hidden variables of the recurrent network can be written as a general non-linear function  $\*h_j = f_{\theta}(\*h_{j-1},\*a_{j})\, \label{eq:arrivalstransition}$, where $\theta$ denotes the network's parameters. Performing the integration in \Eqref{eq:point_likelihood} one obtains

\begin{eqnarray*}
\raggedleft
f^*(t) & = & \exp\left\{\*v^t \cdot \*h_j + w^t\left(t-a_j\right) + b^t + \frac{1}{w^t}\exp\left\{\*v^t\cdot \*h_j + b^t\right\} \right. \\ 
 & & \left. - \frac{1}{w^t}\exp\{\*v^t\cdot\*h_j + w^t\left(t-a_j\right) + b^t\} \right\}.
\end{eqnarray*}
We can learn the model parameters by maximizing the joint model log-likelihood $\mathcal{L}_{\mbox{RPP}} = \sum_{i}\log f^*(\delta_{i+1}|\*h_i)$, where $\delta_{i+1} = a_{i+1}-a_i$ denotes the inter-arrival time.

\subsection{Prediction and Sampling} 
We start by denoting $P(T|\mathcal{H}_j)$ as the distribution that the next point arrives at $T$ given the previous history until $a_j$ - we require $P(T|\mathcal{H}_j)$ for both prediction and sampling. First, notice that the probability of no point arriving between $a_j$ and $a_j + \tau$ can be obtained as an integral over $P(T|\mathcal{H}_j)$, say
\begin{equation*}
\exp\left\{ - \int^{a_j+\tau}_{a_j}\lambda(t)dt\right\} = \int^{\infty}_\tau P(T|\mathcal{H}_j)dT \equiv G(\tau),
\end{equation*}
with $P(T|\mathcal{H}_j) = -\frac{dG(T)}{dT}$, where we used the Poisson distribution for zero arrivals in the first expression.  Solving for $G(\tau)$ we find
\begin{equation}
G(\tau) = \exp\left\{-e^{\alpha_j}\frac{1}{w^t}\left(e^{w^t\tau}-1 \right) \right\},
\label{eq:g-function}
\end{equation}
with $\alpha_j = \*v^t\*h_j + b^t$. The average time of the next arrival is then given by
\begin{equation*}
\mathbb{E}[T] = \int^\infty_0P(T|\mathcal{H}_j)T \, dT = \int^\infty_0 G(T) \, dT.
\end{equation*}
Finally, in order to sample the next arrival time one can use inverse transform sampling on $P(T)$. To this end one requires the inverse of the cumulative function of $P(T)$. We calculate the cumulative function thus
\begin{equation*}
F[P(T|\mathcal{H}_j)] = \int^\tau_0P(T|\mathcal{H}_j)dT = -\int^\tau_0\frac{dG(T)}{dT}dT = G(0) - G(\tau) 
\end{equation*}
whose inverse function then follows
\begin{equation*}
F^{-1}[P(T|\mathcal{H}_j)](y) = \frac{1}{w^t} \left(- \alpha_j + \log{\left\{w^t \left(\log{\left\{ \frac{-1}{y - 1} \right\}} + \frac{e^{\alpha_j}}{w^t}\right) \right \}}\right).
\end{equation*}

\begin{figure}
    \centering
    \includegraphics[width=5.7cm]{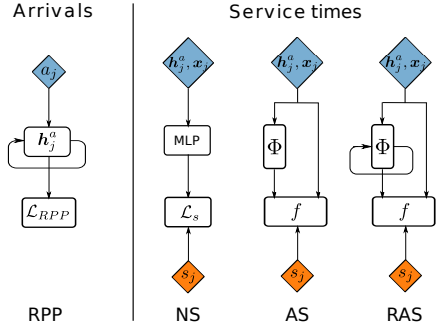}
    \caption{Deep service time models. Left panel: the customer arrivals $a_j$ are modeled using RPP with hidden state $\*h_j$. Right panel: the service time models take the hidden state of the arrival model $\*h_j$ and covariates $\*x_j$ as input and infer the service time distribution.}
    \label{fig:models}
\end{figure}

\section{Models: Deep Service Times}
\label{sec: DeepServiceTimes}
In this Section we introduce our models. We start by setting the methodology with some notation common across our different contributions. Consider the $G/G/\infty$ queuing system. For a given observation time window $[0,\, T]$, where $T$ is the maximum observation time, the dataset consists of a series of arrivals $a_i$ and associated departures $d_i$ each represented in continuous time $\mathbb{R}^+$. Additionally, for each arrival we also have a set of covariates $\*x_i$, upon which we will condition our service time models-these covariates can be taken as scalars or vectors. Now, if for a given $a_i$ no departure is observed within the observation window we set $d_i = \infty$. Thus we define $\mathcal{D}$ as the set of \textit{uncensored events}, i.e. the set of arrivals which have departure within our observation time. Accordingly, $\mathcal{C}$ as the set of \textit{censored events} which have no departure within this window.

Fig. \ref{fig:models} shows an overview of our service time distribution models, which take the hidden state of a \textit{trained} RPP model as input thus providing a rich representation of the customer arrival dynamics. We provide two methodologies: 
\textbf{(i)} we propose parametric forms for the service time distribution where the parameters of known service distributions are defined by multilayered perceptrons. We shall refer to these model as neural service (\textbf{NS-X}) models, where the \textbf{X} labels an specific survival distributions (see Section \ref{NS}). This set of models is a natural generalization of classical stationary solutions to the queuing problem \cite{Kingman} and will serve as our baseline models in what follows;
\textbf{(ii)} we propose two adversarial solutions: first, a static one in which the dynamics of the system is encoded only through the arrivals RPP process (\textbf{AS} model). Second, a dynamical model, wherein the adversarial generator encodes the dynamics of the  systems service via a non parametric state transition function parametrized by a recurrent neural networks (\textbf{RAS} model). 

We will discuss each of these models in detail in Sections \ref{NS}-\ref{RAS}. Finally in Section \ref{mempool models} we focus on a specific problem: the mempool of unconfirmed transactions in the Bitcoin Blockchain transactions network. The dataset for this problem differs from the classical customer arrival process, which instead of being a point process is given as a counting process --- the number of unconfirmed Bitcoin transactions.

\subsection{Neural Service Times (NS-X)}\label{NS}

We start with the customer arrival point process $a_i$. We consider the RPP model as defined in \Eqref{eq:intensity} and denote its hidden state representation as $\*h_j^a = g_{\eta}(\*a_j, \*h_{j-1}^a)$, where $\eta$ labels the set of parameters of the RPP network.

To model the distribution of customer service times, we introduce the generative model
\begin{equation*}
s_i \sim \Phi_{\theta}(s|\*h^a_i,\*x_i),
\end{equation*}
with parameter set $\theta$. This model  captures the complicated dependencies in the arrival dynamics $a_i$ --- encoded through the hidden states $\*h^a_i$, and any other covariates $\*x_i$ in the system. This conditional form allows our model to leverage the dynamical information of the arrival process. We define $\Phi_{\theta}$ as one of the following five distributions: \textit{Gamma} (NS-G), \textit{Exponential} (NS-E), \textit{Pareto} (\textbf{NS-P}), \textit{Chi-square} (\textbf{NS-C}) or \textit{Log-normal} (\textbf{NS-L}), whose parameter set $\mathcal{P}$ are defined via multilayer perceptrons. Thus for the \textbf{NS-G} model we have $s_i \sim \mbox{Gamma}(\alpha^a_i,\beta^a_i)$ with $\mathcal{P} = \left [\alpha^a_i, \beta^a_i \right]=\mbox{MLP}_{\theta}([\*h^{a}_i,\*x_i])$. The neural service models can then be interpreted as a marked RPP where the marks are continuous and have a dynamical character whose distribution corresponds to that of the service times. We train these models by maximizing the log-likelihood of our generated service times with respect to the uncensored dataset $\mathcal{D}$.

 \textbf{Censored events.} To capture censored events we introduce the probability of obtaining an expected remaining service time bigger than the observation window $T_i = T - a_i$
 \begin{equation*}
     \bar{\Phi}(s_i) = \int^{\infty}_{T_i}\Phi_\theta\left(\tau|,\*h^a_i,\*x_i\right)d\tau \, .
 \end{equation*}
 The complete log-likelihood of the \textbf{NS-X} model then reads
\begin{equation*}
\mathcal{L}_{s} = \sum_{\mathcal{D}}\log\left\{\Phi_\theta\left(s_i|\*h^a_i,\*x_i\right)\right\} + \sum_{\mathcal{C}}\log\left\{\bar{\Phi}\left(s_i\right)\right\}.
\end{equation*}

The \textbf{NS-X} models are provided with output distributions which are common solutions to  stationary distributions for service times in theoretical models \cite{asmussen2008applied}. The NS models are the closest representation to these theoretical estimates, due to the distributional forms of the outputs distribution. Queuing theoretical models, however, cannot capture the censored arrivals, in contrast to the NS-X models. Therefore the NS-X models are a natural extension of the theoretical ones and serves as a baseline to the adversarial models introduced in the next section.

\begin{figure*}[h!]
     \begin{subfigure}{0.36\textwidth}
        \centering
        \includegraphics[width=\textwidth]{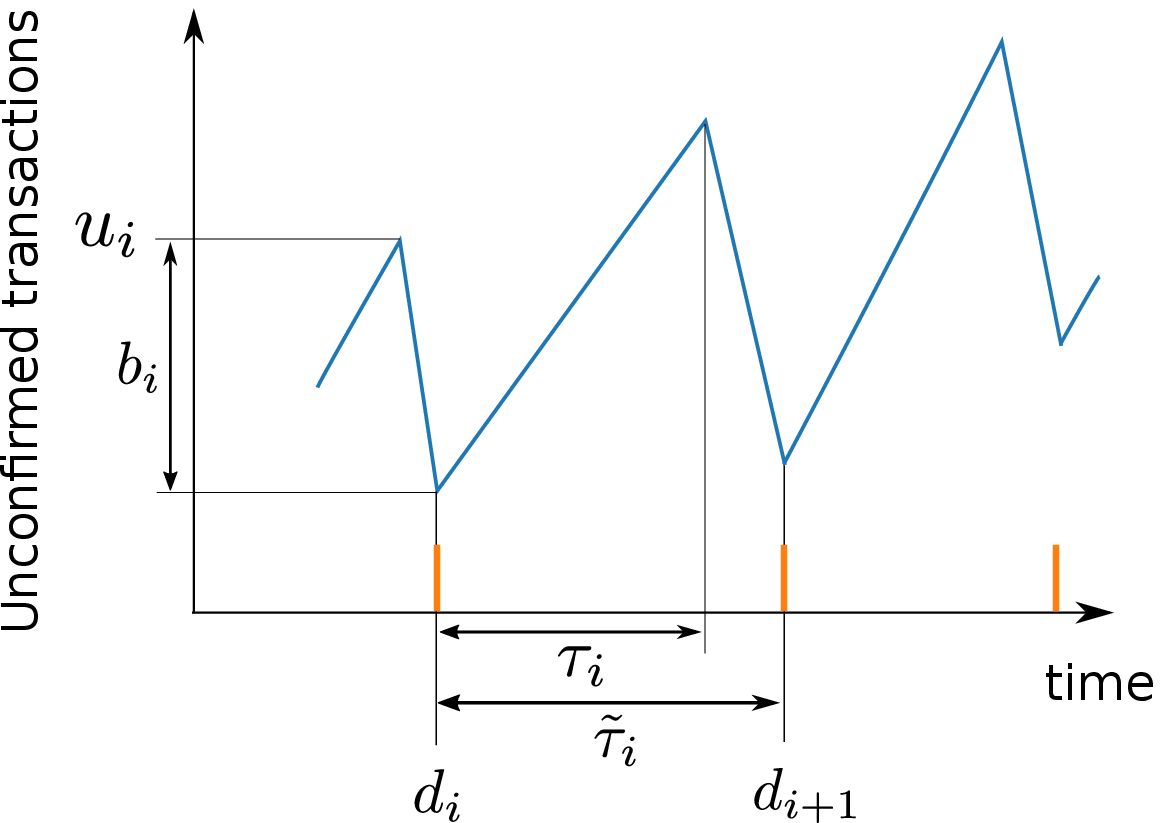}
        \caption{Raw Mempool Data}
        \label{fig:mempool_raw}
    \end{subfigure}%
    \hspace{0.7cm}
    \centering
    \begin{subfigure}{0.46\textwidth}
        \centering
        \includegraphics[width=\textwidth]{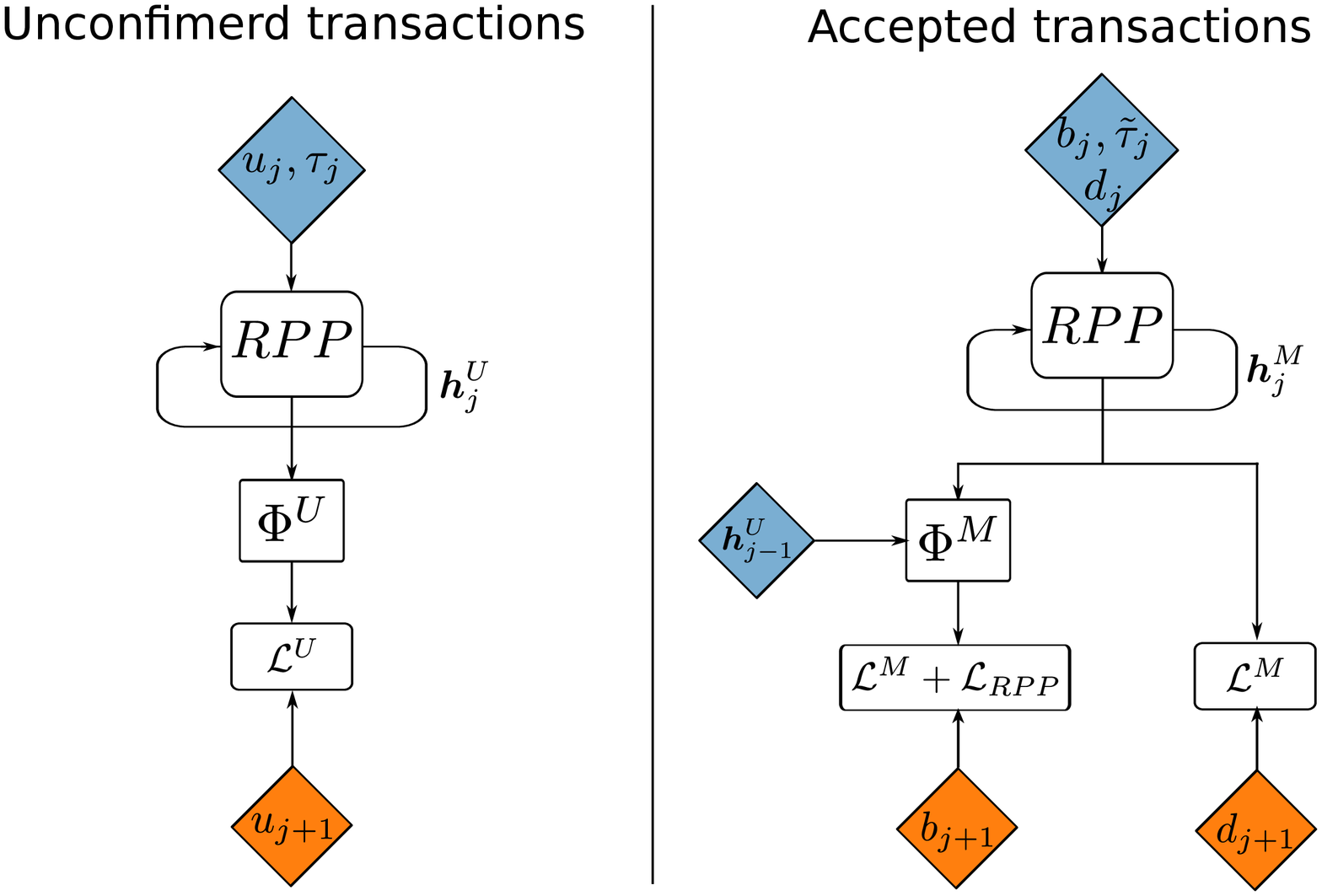}
        \caption{Mempool models}
        \label{fig:mempool_models}
    \end{subfigure}%
     \caption{Mempool Specifics. Panel (a) depicts the raw Mempool data, $u_i$ number of unconfirmed transactions before block creation $d_i$, and $b_i$ is the number of transactions in the block $d_i$. Deep service time models for the Mempool data. Left panel (b): the number of unconfirmed transactions $u_j$ are modelled using RPP with hidden state $\*h^a_j$. Right  panel (b): \textit{block creation} is modeled using and RPP with hidden state $\*h^U_j$. The accepted transactions models take the hidden state of the unconfirmed transactions $\*h^U_j$, the hidden state of the block creation $\*h^U_j$ and covariates $\*x_j$ as input and infer the size of the block (number of accepted transactions).} 
    \label{fig:mempool_theory}
\end{figure*}

\subsection{Adversarial Service Times (AS)}\label{AS}
The expressibility of our model is severely constrained by any functional form imposed on $\Phi(s)$. In order to allow for more general service time distributions we consider Generative Adversarial Networks (GAN)~\cite{NIPS2014_5423}, a class of generative models wherein approximate (service time) samples $s_i$ are drawn as
\begin{equation*}
s_i  = \Phi_{\theta}(s | \epsilon, \*h^a_i, \*x_i), \hspace{0.5cm} \epsilon \sim \mathbb{P}_{\epsilon}\mbox{.}   
\end{equation*}

Here $\mathbb{P}_{\epsilon}$ is a simple distribution, e.g. isotropic Gaussian or uniform distribution, and the \textit{generator} $\Phi_{\theta}$ is modeled by a deep neural network with parameter set $\theta$, conditioned on both the arrival dynamics and the system's covariates. In our experiments we define $\Phi_{\theta}$ as a multilayer \footnote{In our experiments, we found that 3 layers provides the best model} perceptron and add a noise term drawn from a Gaussian $\mathcal{N}(0, 1)$ at each of this layers, as to increase the variance in the samples from $\Phi_{\theta}$ \cite{Chapfuwa2018AdversarialTM}.

This class of models is trained by minimizing specific distances (or divergences) between the empirical distribution --- here the distribution of \textit{uncensored} events $\mathbb{P}_{\mathcal{D}}$, and the distribution $\mathbb{P}_{\theta}$ of the generated samples $\{ \Phi_{\theta}(s) \}$. Each such distance differs on the impact it has on the convergence of $\mathbb{P}_{\theta}$ towards the empirical distribution, and thus on the training stability. Here we choose to minimize the Wasserstein-1 distance (WGAN) \cite{pmlr-v70-arjovsky17a}, which has been shown to be continuous everywhere and differentiable almost everywhere, as opposed to e.g. the Jensen-Shannon divergence minimized in the original GAN formulation. 

Using the Kantorovich-Rubinstein duality \cite{villai_transport} to compute the Wasserstein-1 distance one can express the WGAN objective function $\mathcal{L}$ as:
\begin{equation}
    \mathcal{L} = \underset{\theta}{\mbox{min}} \underset{f_{\varphi} \in \mathfrak{L}_1}{\mbox{max}} \mathbb{E}_{s \sim \mathbb{P}_{\mathcal{D}}} \left[ f_{\varphi}(s)\right] - \mathbb{E}_{s \sim \mathbb{P}_{\theta}} \left[ f_{\varphi}(s)\right],
    \label{eq:wgan_loss}
\end{equation}
where the maximum is taken over all 1-Lipschitz functions $\mathfrak{L}_1$, defined as functions whose gradients have norms at most 1 everywhere (all the functions we consider are sufficiently smooth). Within the WGAN formulation the critic function $f_{\varphi}$ is modeled by a deep neural network with parameter set $\varphi$ and needs to fulfill the Lipschitz constraint. In order to enforce it we follow~\cite{petzka2018on} and add a regularization term of the form 
\begin{equation*}
    \mathcal{L}_1 = \mathbb{E}_{s \sim \mathbb{P}_i} \left[\left( \mbox{max} \left\{0, |\nabla_{s} f_{\varphi}(s)|-1 \right\} \right)^2 \right],
    \label{eq:lipschitz}
\end{equation*}
where $\mathbb{P}_{i}$ is implicitly defined as sampling uniformly along straight lines between pairs of points sampled from the empirical $\mathbb{P}_{\mathcal{D}}$ and the generator $\mathbb{P}_{\theta}$ distributions \cite{Gulrajani2017ImprovedTO}. Minimizing \Eqref{eq:wgan_loss} under an optimal critic function with respect to $\theta$ minimizes the Wasserstein distance between $\mathbb{P}_{\mathcal{D}}$ and $\mathbb{P}_{\theta}$ --- this defines the adversarial game.

In our experiments the critic is defined as a 3-layer perceptron and is also conditioned on the covariates $f_{\varphi} = f_{\varphi}(s, \*x)$.

\textbf{Censored events.} To train $\Phi_{\theta}$ to learn the distribution of censored events $\mathbb{P}_\mathcal{C}$ we follow \cite{Chapfuwa2018AdversarialTM} and consider a second regularizer which penalizes sampled service times smaller than the censoring time $T$, that is
\begin{equation*}
\mathcal{L}_2 = \mathbb{E}_{\*x \sim \mathbb{P}_{\mathcal{C}}, \epsilon \sim \mathbb{P}_{\epsilon}} \left[\mbox{max}\left\{ 0, T-\Phi_{\theta}(s|\epsilon, \*h^a, \*x) \right\} \right].
\end{equation*}
We also correct for situations in which the proportions of uncensored events is low through
\begin{equation*}
\mathcal{L}_3 = \mathbb{E}_{\epsilon \sim \mathbb{P}_{\epsilon}, (\tilde{s}, \*x) \sim \mathbb{P}_{\mathcal{D}}} \left[\,\left|\tilde{s}-\Phi_{\theta}(s|\epsilon, \*h^a, \*x)\right|\,\right].
\end{equation*}

Our full objective function reads $\tilde{\mathcal{L}} = \mathcal{L} + \sum\limits_{i=1}^3 \lambda_i \mathcal{L}_i$, where $\lambda_1 = 10$ and $\lambda_3 = 1$ throughout all experiments whereas $\lambda_2$ changes depending on the datasets (see Section \ref{subsec:training_dets} for details).

\subsection{Recurrent Adversarial Service Time (RAS)}\label{RAS}
The response of a service system to newly arrived customers intuitively has a dynamic component (e.g. the dynamic reallocation of resources depending on the number of arrivals still on service, the response to past events disrupting the service, etc). In order to capture such a dynamic response and implicitly characterize exogenous events we approximate the system's transition function with a \textit{stochastic} recurrent neural network 
\begin{equation}
\*h_i^{\Phi} = g_{\theta}(\epsilon_i, \*h_i^a, \*x_i, \*h_{i-1}^{\Phi}), \hspace{0.5cm} \epsilon_i \sim \mathbb{P}_{\epsilon},
\label{eq:rnn-gan}
\end{equation}
where $g_{\theta}$ is a RNN with parameter set $\theta$ and $\*h_i^{\Phi}$ is the hidden state encoding the independent dynamic character of the service system. The model is informed about the incoming arrival through the hidden state $\*h_i^a$ of the arrival RPP model and the arrival covariates $\*x_i$. Its noisy component, on the other hand, comes from $\mathbb{P}_{\epsilon}$, an isotropic Gaussian sampled at each arrival time.

We then define the generator $s_{i} = \Phi_{\theta}(s| \*h_i^{\Phi}, \epsilon), \epsilon\sim\mathbb{P}_\epsilon$ as a 3-layer perceptron with the RNN's hidden representation \Eqref{eq:rnn-gan} as input and an additional noise terms $\epsilon$ added in each layer. We train the model by minimizing \Eqref{eq:wgan_loss} together with the regularizers $\mathcal{L}_i$ as above. Let us note here that recurrent generative models with adversarial training has been considered before \cite{Mogren2016CRNNGANCR}, \cite{hyland2018realvalued}. In our experiments the critic function $f_{\varphi} = f_{\varphi}(s, \*x_i)$ remains static and is defined once more as a 3-layer perceptron.

\subsection{Bitcoin Mempool}\label{mempool models}
In the following we modify our approach to analyze data from a specific type of queuing: the transaction queuing of the Bitcoin network. The decentralized currency protocol known as Bitcoin \cite{nakamoto2008bitcoin} utilizes a peer-to-peer (P2P) architecture that enables users to send and receive transactions, denominated in units of Bitcoin (BTC).  Transactions are broadcasted by a Bitcoin client and received by the peer-to-peer network. They are confirmed after having been added to the Blockchain. This data structure contains blocks of all accepted transactions since the genesis of the system. The creation of each block defines a point process which can be understood as a departure process for the transactions, thus encouraging the modelling of the system dynamics as a queuing system. Specifically, we analyze the Bitcoin \textit{mempool}, the set of unconfirmed transactions $u$ in the Bitcoin network. Here, the creation of a block at time $d_i$ generates a sudden drop $b_i$ in the number of unconfirmed transactions $u_i$ (Fig. \ref{fig:mempool_raw}). The set of unconfirmed transactions plays the role of \textit{waiting clients}, and the creation of a block specifies the simultaneous \textit{departure} of many clients (transactions). 

We first model the independent dynamics of the number of unconfirmed transaction with the generative model
\begin{equation}
    u_{i+1} = \Phi_{\theta}^U(u | \*h^U_i), \hspace{0.5cm} \*h^U_i = g_{\theta} (\epsilon, \tau_i, u_i, \*h^U_{i-1}).
    \label{u-model-meempool}
\end{equation}
Here $g_{\theta}$ is a RNN with parameters $\theta$, $\*h^U_i$ encodes the history of $u_i$ and $\tau_i$ is shown in Fig. \ref{fig:mempool_raw}. We present two approximations to the mempool problem: (i) a parametric one, which we called Neural Meempool Service (\textbf{NMS-G}) for which $\Phi_{\theta}^U = \mbox{Gamma}(\alpha^u_i,\beta^u_i)$ with $\left [\alpha^a_i, \beta^a_i \right]=\mbox{MLP}_{\theta}(\*h^{U}_i)$ and $\epsilon = 0$; (ii) and a nonparametric one, which we called Adversarial Mempool Service (\textbf{AMS}), in which $\Phi_{\theta}^U$ is given by a 3-layer perceptron and for which $\*h^U_i$ is a now random variable with $\epsilon \sim \mathbb{P}_\epsilon = \mathcal{N}(0, 1)$ in \Eqref{u-model-meempool}.

Now, the creation of the blocks which form part of the Blockchain defines a departure process which we describe using a RPP model with intensity function
\begin{equation*}
    \lambda_d^{*}\left(t\right)=\exp{\{\mathbf{v}^t_M\cdot\mathbf{h}_j^M+\mathbf{v}_U^t\cdot\mathbf{h}_j^U+w^t\left(t-d_j\right)+b^t\}},
\end{equation*}
where $\*h^U_i$ contains the dynamic information of the unconfirmed transaction process whereas $\*h^M_i = \tilde{g}_{\phi} (b_i, d_i, \tilde{\tau}_i, \*h^M_{i-1})$, with $\tilde{g}_{\phi}$ a RNN describing the departure dynamics. Finally we introduce a generative model for the accepted transaction thus
\begin{equation*}
    b_{i+1} = \Phi^M_{\phi}(b | \epsilon, \*h^M_i, \*h^U_{i-1}), 
\end{equation*}
with $\epsilon=0$ and $\Phi^M_{\phi}$ a Gamma function in our \textbf{NMS-G} formulation, or $\epsilon \sim \mathcal{N}(0, 1)$ and $\Phi^M_{\phi}$ a 3-layer perceptron in our nonparametric \textbf{AMS} version. We train the \textbf{NMS-G} model via maximum likelihood and \textbf{AMS} using Eqs. \ref{eq:wgan_loss} and \ref{eq:lipschitz}; the RPP block-creation model is trained as described in Section \ref{ssec:RMPP}. The complete overview of the mempool models is given in Fig. \ref{fig:mempool_models}.
\begin{figure*}[t!]
    \centering
    \begin{subfigure}{0.33\textwidth}
        \centering
        \includegraphics[width=6.cm, keepaspectratio]{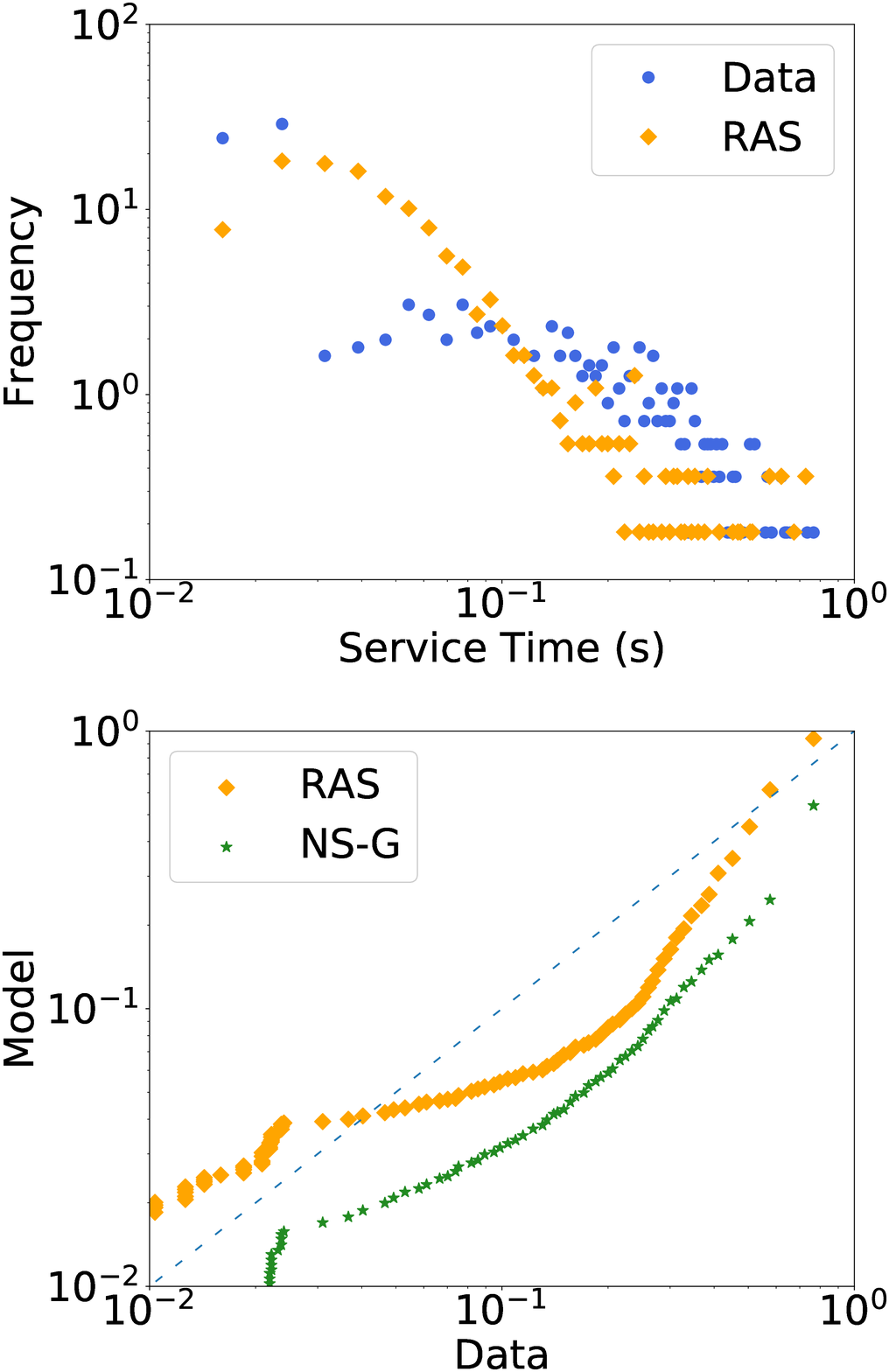}
        \caption{NH-PT}
    \end{subfigure}%
    \begin{subfigure}{0.33\textwidth}
        \centering
        \includegraphics[width=6.cm, keepaspectratio]{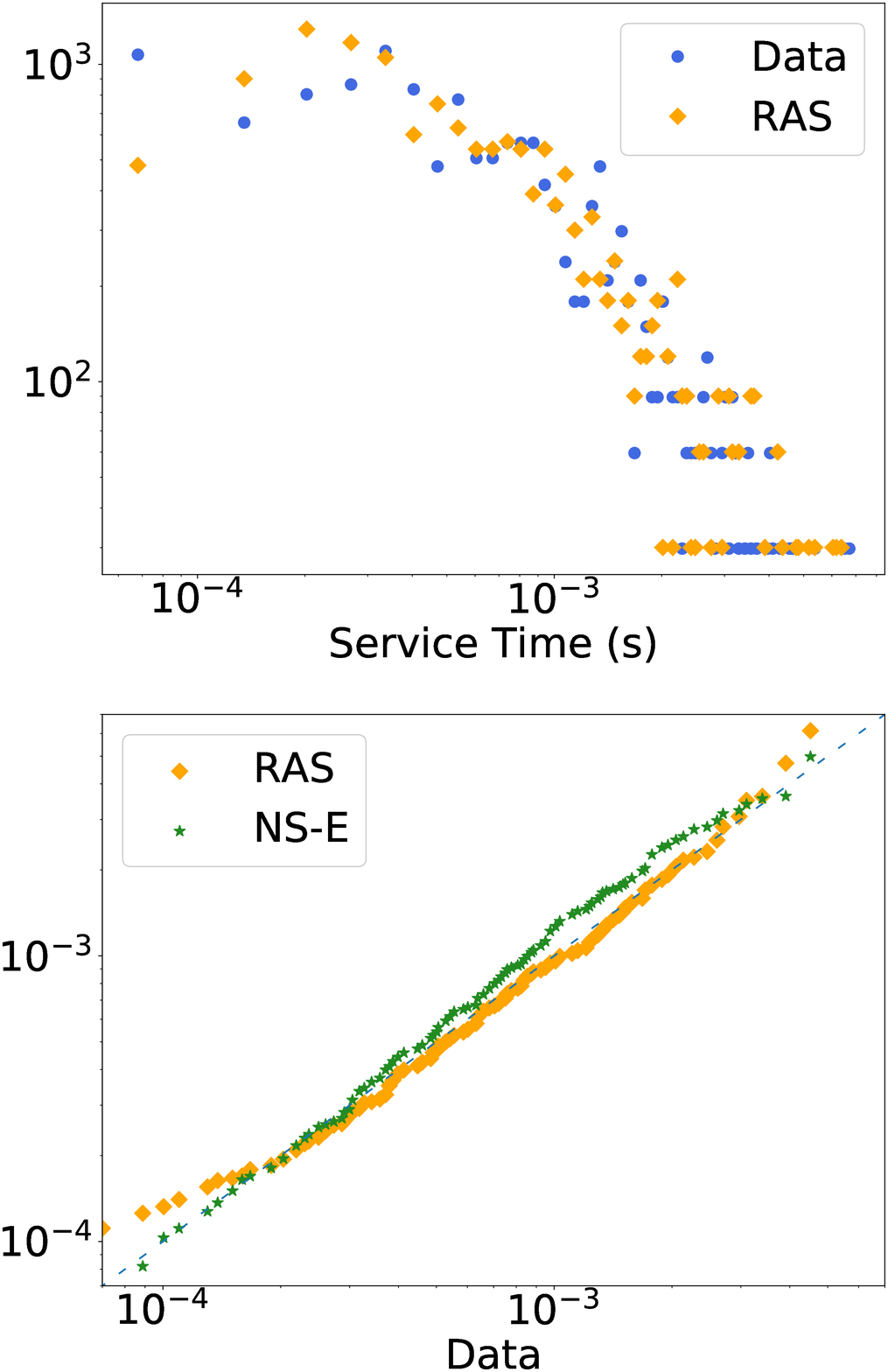}
        \caption{NH-PS}
    \end{subfigure}
    \begin{subfigure}{0.33\textwidth}
        \centering
        \includegraphics[width=6.cm, keepaspectratio]{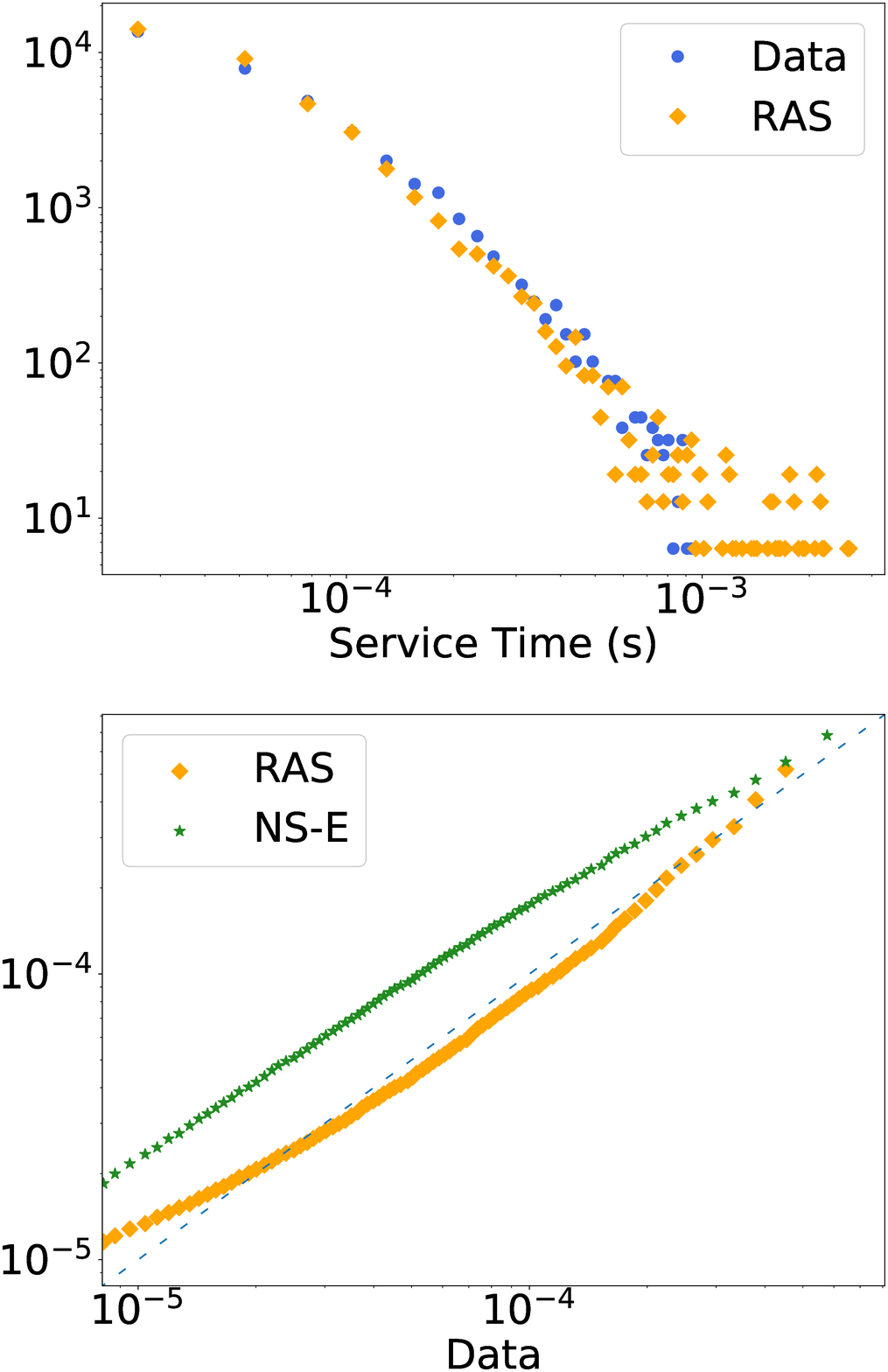}
        \caption{H-PS}
    \end{subfigure}
    \caption{Comparison between the synthetic simulated data distribution and our best model. Q-Q plots against empirical distribution for both the best theoretical neural model and our best adversarial solution.}
    \label{fig:distributions-synthetic}
\end{figure*}

\begin{table*}[t!]
\caption{Main results on synthetic datasets.}
\begin{center}
\setlength{\tabcolsep}{0.7em}
\begin{tabular}{lcccccccc}
  {}    & \multicolumn{2}{c}{NH-PT}      & \multicolumn{2}{c}{NH-PS}     & \multicolumn{2}{c}{H-PT}    & \multicolumn{2}{c}{H-PS}       \\
  \hline
  {mean} & \multicolumn{2}{c}{0.052}     & \multicolumn{2}{c}{0.0004}    & \multicolumn{2}{c}{0.0061} & \multicolumn{2}{c}{2.125e-5}  \\
  {}   & error              & KS               & error      & KS               & error           & KS      & error            & KS     \\
  \hline
  NS-G   & \textbf{0.207} & 0.218          & 0.020           & 0.520          & 0.254         & 0.879      & \textbf{2.18e-5} & 0.401  \\
  NS-E   & 0.209          & 0.154          & 0.0006          & 0.082          & 0.432         & 0.979      & 3.06e-5          & 0.501  \\
  NS-P   & 0.210          & 0.330          & 0.0007          & 0.610          & 0.443         & 0.981      & 2.09e-5          & 0.501  \\
  NS-C   & 0.372          & 0.242          & 0.061           & 0.527          & 0.037         & 0.988      & 0.029            & 0.511  \\
  NS-L   & 5.293          & 0.525          & 6.158           & 0.479          & 4.282         & 0.971      & 8.500            & 0.555  \\
  \hline
  AS     & 0.215          & 0.113          & 0.0016          & 0.448          & 0.250         & 0.235      & 1.24e-4          & 0.121  \\
  RAS-NH & 0.218          & 0.124          & 0.0031          & 0.062          & 0.249         & 0.222      & 1.37e-4          & 0.136  \\
  RAS    & \textbf{0.207} & \textbf{0.094} & \textbf{0.0005} & \textbf{0.042} & \textbf{0.242}& \textbf{0.212}& 1.09e-4       & \textbf{0.110}
\end{tabular}
\end{center}
\label{tab:synthetic_results_table}
\end{table*}
\section{Experiments} \label{sec: experiments}
\begin{figure*}[t!]
    \centering
    \begin{subfigure}{0.33\textwidth}
        \centering
        \includegraphics[width=6.cm, keepaspectratio]{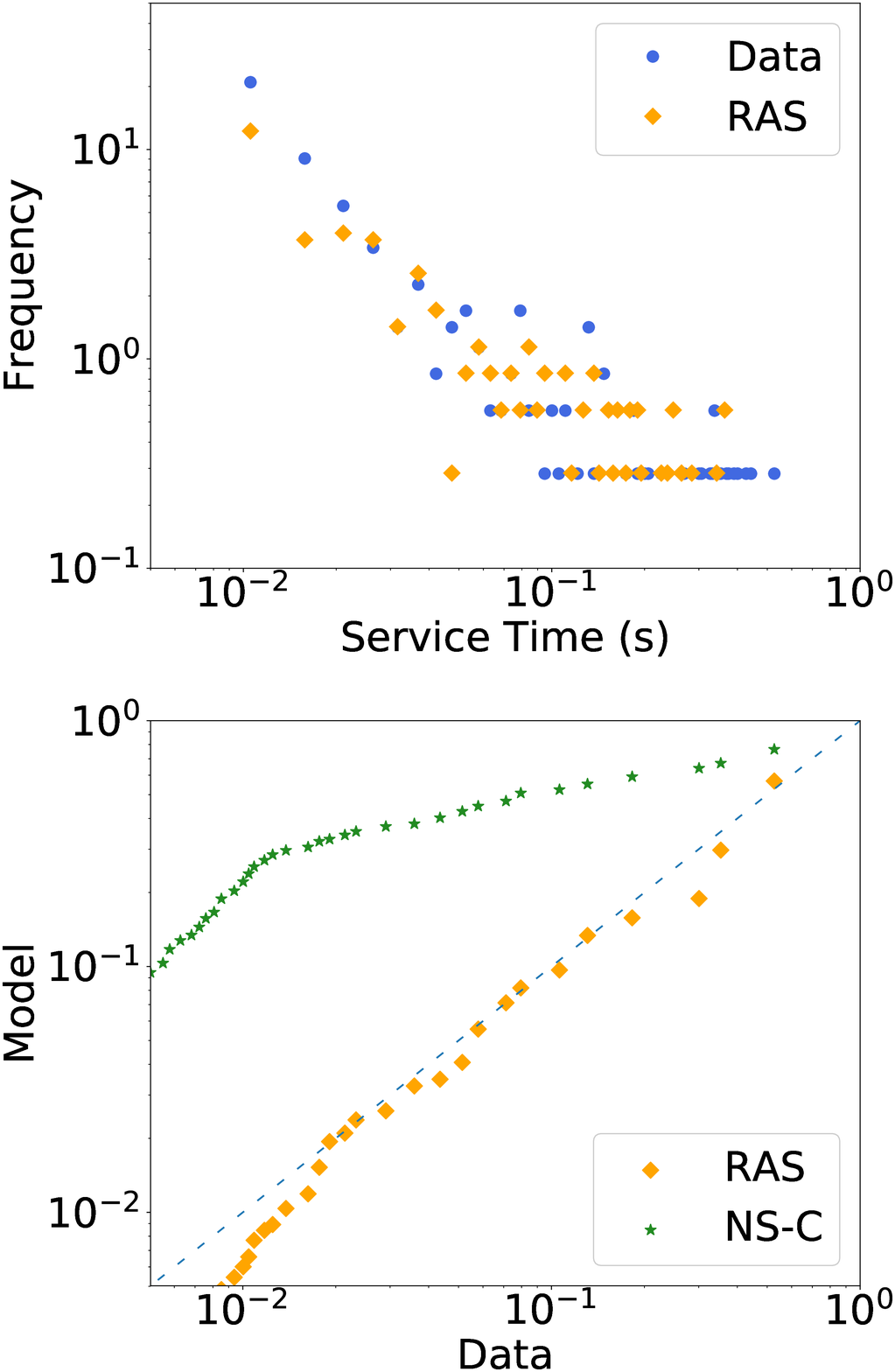}
        \caption{Github}
    \end{subfigure}%
    \begin{subfigure}{0.33\textwidth}
        \centering
        \includegraphics[width=6.cm, keepaspectratio]{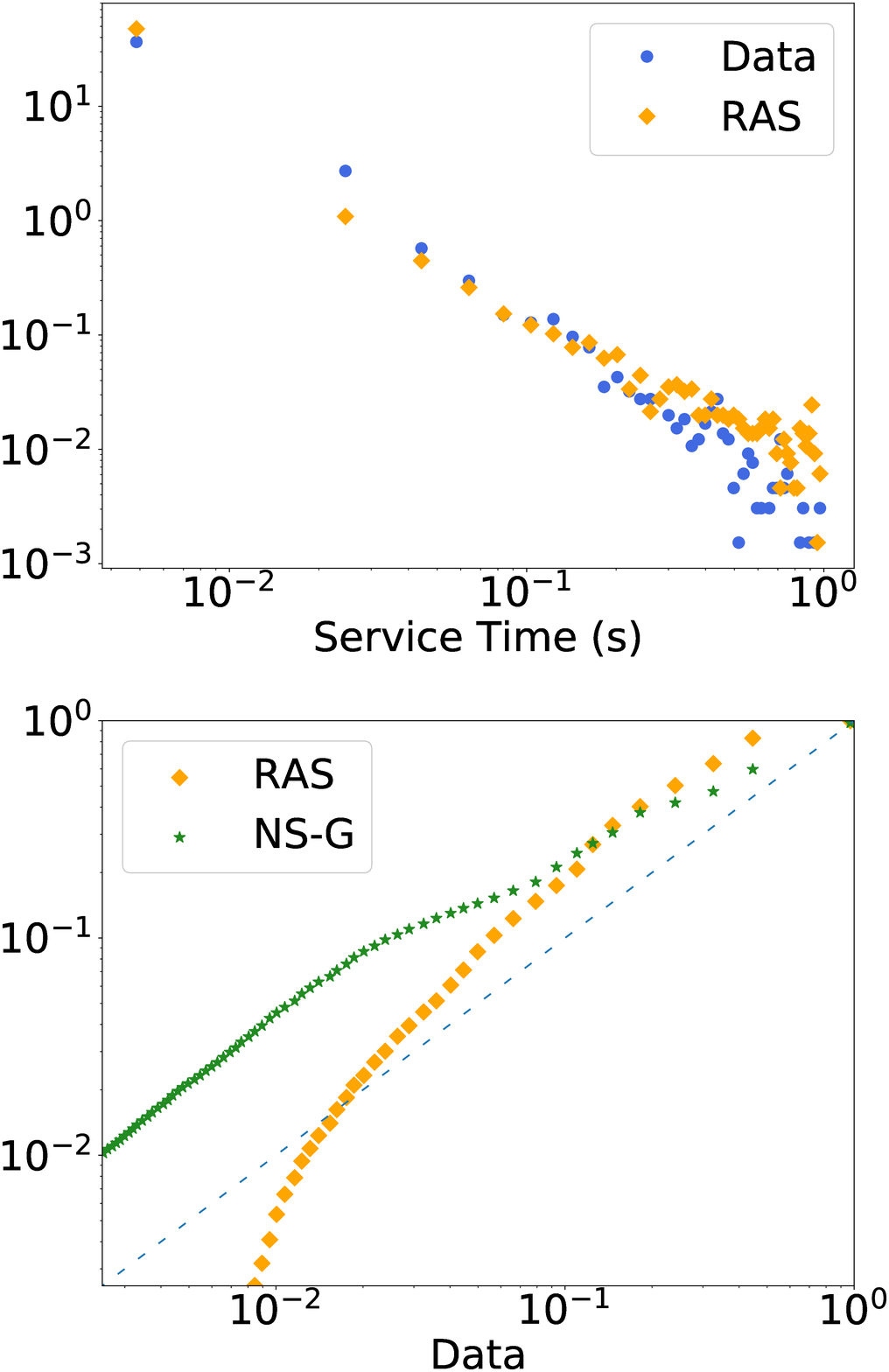}
        \caption{Stackoverflow}
    \end{subfigure}
    \begin{subfigure}{0.33\textwidth}
        \centering
        \includegraphics[width=6.cm, keepaspectratio]{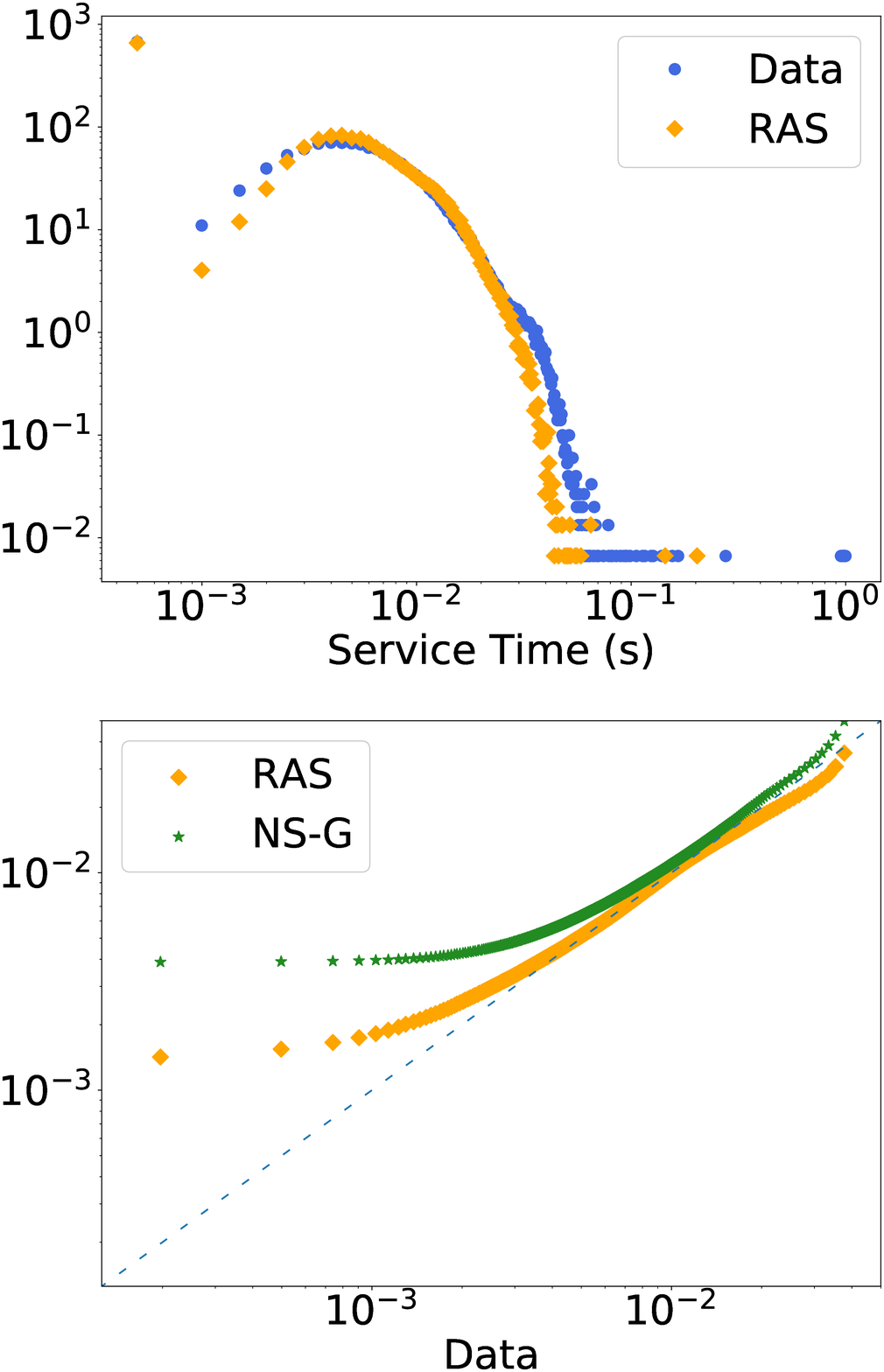}
        \caption{NY}
    \end{subfigure}
    \caption{Comparison between the empirical data distribution and our best model. Q-Q plots against empirical distribution for both the best theoretical neural model and our best adversarial solution.}
    \label{fig:empirical_service_distributions}
\end{figure*}
\begin{table}
\caption{Main results on empirical data-sets. }
\begin{center}
\setlength{\tabcolsep}{0.7em}
\begin{tabular}{lcccccc}
  {}         &  \multicolumn{2}{c}{Github}          & \multicolumn{2}{c}{NY}          & \multicolumn{2}{c}{Stackoverflow}\\
  \hline
  {mean} &   \multicolumn{2}{c}{0.0113}            & \multicolumn{2}{c}{0.0068}         & \multicolumn{2}{c}{0.0193} \\
  {}                  & error     & KS                     & error       & KS                      & error     & KS \\
  \hline
  NS-G             & 0.073          & 0.396         & 0.025         & 0.154                    & 0.378         & 0.480 \\
  NS-E             & 0.074          & 0.458         & 0.007         & 0.251                    & 0.379         & 0.509 \\
  NS-P             & 0.071          & 0.604         & 0.008         & 0.367                    & 0.378         & 0.466 \\
  NS-C             & 0.096          & 0.341         & 0.182         & 0.632                    & 0.403         & 0.533 \\
  NS-L             & 6.37           & 0.496         & 0.155         & 0.627                    & 15.93         & 0.595 \\
  \hline
  AS               & \textbf{0.071} & 0.039         & 0.006         & 0.094                    & 0.383         & \textbf{0.226} \\
  RAS-NH           & 0.112          & 0.240         & 0.098         & 0.165                    & 0.388         & 0.492 \\
  RAS              & 0.072          & \textbf{0.034}& \textbf{0.005}& \textbf{0.030}           & \textbf{0.369}& 0.281
\end{tabular}
\end{center}
\label{tab:empirical_results_table}
\end{table}

In this Section we provide the experimental framework upon which we tested our model. First we introduce the datasets which were used in the experiments. We provide synthetic datasets with established models for both the arrivals and the service processes, as well as empirical datasets - this demonstrates the ability of our approach to handle diverse application areas in an flexible and scalable manner. Finally, we specify the details of the neural networks architectures implemented for the experiments, as well as learning parameters and any other hyperparameters as required in the model specification.

\subsection{Synthetic datasets}
 In order to provide a controlled environment to test the behavior of our methodology we introduce the following datasets for different arrivals and service processes. We consider two different arrival processes:
 (i) \textbf{Hawkes Process (HP)} \cite{hawkes1974cluster}, which is a model for self-exciting phenomena where user arrivals increase the probability of other users to arrive. It is widely used as a model for users on internet services as it is able to capture bursty human dynamics and is defined via the following conditional intensity function $\lambda_H(t) = \lambda_{0} +\sum_{T_i:t > T_i}\mu(t - T_i)$,  where $\lambda_{0}$ corresponds to the base intensity and models the exogenous arrival events, and $\mu(t - T_i)$ is the memory kernel which provides the intensity given by past arrivals $T_i$.
(ii) \textbf{Non-linear Hawkes Process}: an extension of the Hawkes process that allows for inhibitory behavior through a non-linear function over the history of arrivals $\lambda_{NH}(t) = \phi \left(\lambda_H(t)\right)$.

In order to gain intuition into traditional service models, we introduce two service functions definitions:
(i) \textbf{Phase Type Distribution (PT)}: one decomposes the service as a series of exponential service steps. It is defined with the time taken between the initial state and the  absorbing state \footnote{a type of first passage time} in a continuous time Markov chain. It can be shown that such \textit{phase type} distributions can approximate any non-negative continuous distribution.
(ii) \textbf{Processor Sharing Distribution (PS)}: a queuing model in which the system handles an infinite amount of clients simultaneously but must reallocate resources with each new client arrival or departure. It was introduced by Kleinrock \cite{kleinrock1976queueing} to model computer systems that handle multiple client simultaneously. One can think that each client in the system  at any time instantaneously receives $1/U(t)$ service power, where $U(t)$ is the number of unfinished clients (clients in the queuing)\cite{sutton2011bayesian}.

The combinations of the two arrivals and the two service models provide four different synthetic queuing: \textbf{H-PT}, \textbf{H-PS}, \textbf{NH-PT}, \textbf{NH-PS}.

\subsection{Empirical datasets}
We gathered datasets from a variety of internet services, thus providing an in depth analysis of the temporal patterns of users in different domains.\newline
\textbf{Stackoverflow}: a question-answering platform for programmers. We define the customers arrivals as the point in time when questions are posted by the users of the web page. We define the service time as the elapsed time between a question and its subsequent accepted answer time. This view establishes the ensemble of users which provide answers as the service system.  We analyze a total of $2\times10^7$ questions.\newline
\textbf{Github}:  The version control repository and internet hosting service. As customers arrivals, we defined the creation of an issue in a given repository. Its departure time is the moment the given issue was closed. Therefore, the set of users associated with a given repository, can be thought of as the service system. We analysed the top (ranked by the number of issues) 500 repositories in the platform in 2015 in total $1.5\times 10 ^6$ different issues.\newline
\textbf{New York City Taxi Dataset (NY)}: The dataset contains data of individual taxi trips in New York city. Customers arrivals are defined as starting time of the trip and the departure time is defined as the final time of the trip. Here, the service system is provided by both the taxi providing the service and the transportation network of roads, streets and highways pertaining to the city of New York.  \footnote{https://www1.nyc.gov/site/tlc/about/tlc-trip-record-data.page} \newline
\textbf{Mempool}: The dataset consist of all unconfirmed transaction in the mempool dataset as observed by one miner for the time period between January 2017 until June 2018. Here, the service system consists of the whole Bitcoin miner network. This is the only dataset which was used for the Bitcoin mempool model and the details of the arrivals and service is provided in the model specification above.  \footnote{https://jochen-hoenicke.de/queuing}
\begin{figure*}[!ht]
    \centering
    \begin{subfigure}{0.33\textwidth}
        \centering
        \includegraphics[width=6.cm, keepaspectratio]{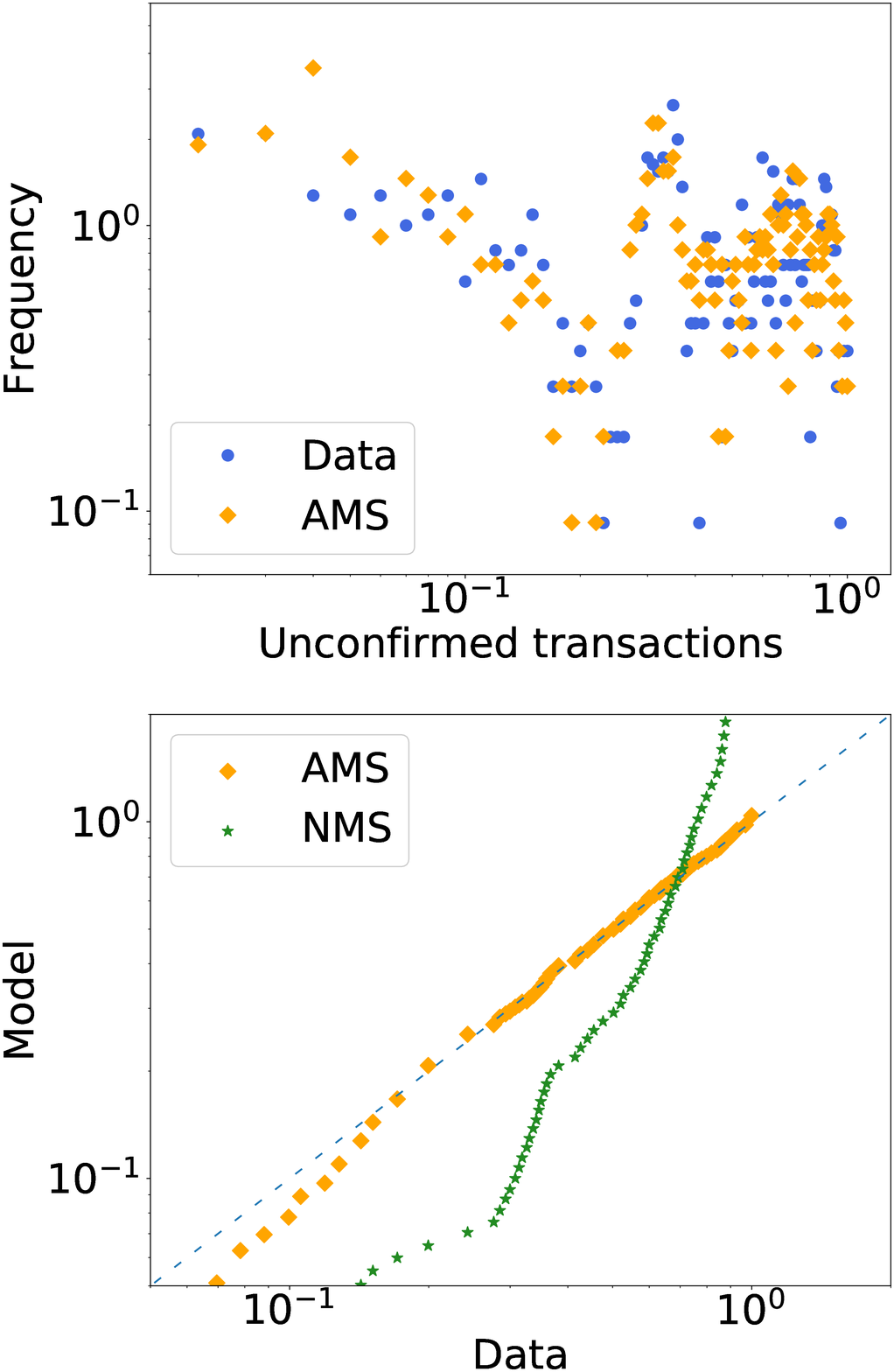}
        \caption{Transactions Adversarial}
    \end{subfigure}
     \begin{subfigure}{0.33\textwidth}
        \centering
        \includegraphics[width=6.cm, keepaspectratio]{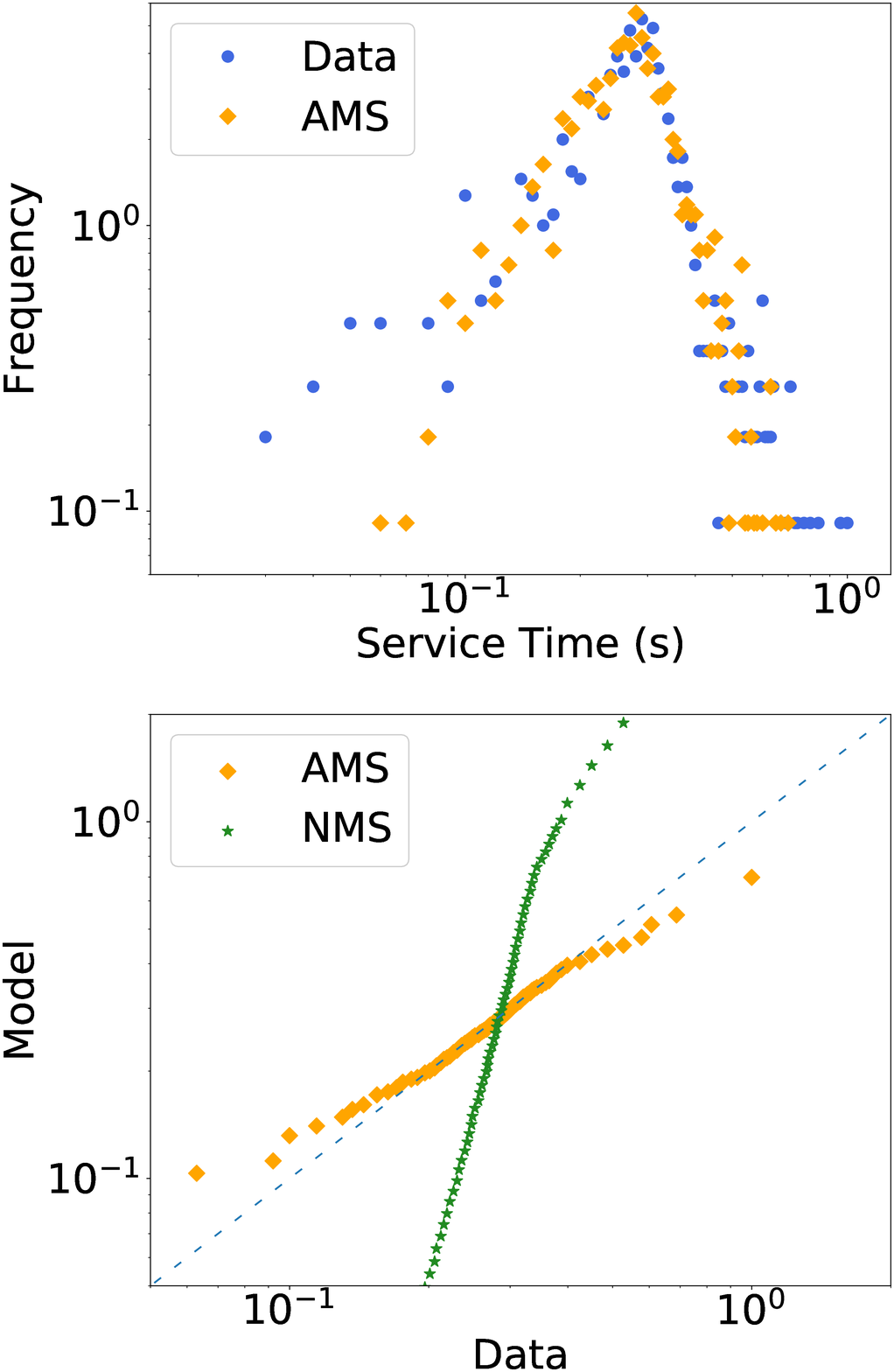}
        \caption{Service Adversarial}
    \end{subfigure}%
    \caption{Comparison of distributions for the mempool dataset. To compare against the baselines, Q-Q plots versus the empirical distribution are also shown.}
    \label{fig:mempool_results}
\end{figure*}
\begin{table}
\caption{Mempool dataset results. }
\begin{center}
\setlength{\tabcolsep}{0.7em}
\begin{tabular}{lcccc}
  {}     & \multicolumn{2}{c}{Unconfirmed Transactions}      & \multicolumn{2}{c}{Departures} \\
  \hline
  {mean} & \multicolumn{2}{c}{0.2484}            & \multicolumn{2}{c}{0.2159}   \\
  {}     & error     & KS                        & error      & KS       \\
  \hline
  P    & 0.2513     & 0.2417                   & 0.1173     & 0.2781    \\
  L    & 0.2687     & 0.2527                   & 0.1173     & 0.4760    \\
  G    & 0.1948     & 0.2550                   & 0.1173     & 0.4760    \\
  \hline
  NMS-G    & 0.1725     & 0.1472                   & 0.3080     & 0.2481    \\
  AMS    & \textbf{0.0271}   &  \textbf{0.0236}                   & \textbf{0.0016}     & \textbf{0.0290} 
\end{tabular}
\end{center}
\label{tab:mempool}
\end{table}
\subsection{Training details}\label{subsec:training_dets}
For the purpose of optimizing the neural networks parameters, we use the ADAM stochastic optimization \cite{ADAM} method with a learning rate of $10^{-4}$ (except for the neural mempool model where we use $10^{-5}$). We split the data into training and test sets. The test set is defined as $\sim 5\%$ of the time series. For all the arrival models, we used the Gated Recurrent Units  \cite{GRU} for the non parametric state transition functions. The dimension of the GRU is 64 for the PS models, 16 for the  PT and Github dataset, 128 for the New York dataset and 256 for the Stackoverflow dataset. For the RAS model, LSTMs where used as state transitions in the arrival model. The MLP for the NS model has two hidden layers with 256 dimensions, whereas the adversarial generator and the critic in the AS and RAS model have 3 hidden layers of 100 units. For the mempool dataset, the NMS-G model requires perceptrons with dimension 32 and LSTM units of size 16 for the unconfirmed transactions. For the service model the dimensions of the perceptrons is 32 and for the LSTM it is 62. The AMS model has 20 units for the unconfirmed transactions generator and critic, and 20 units for the transition function. For the service dynamics, the adversarial model required 10 hidden units for the generator and critic and 10 units for the LSTM transitions.

\section{Results}
\label{sec: discussion}
\begin{figure*}[t!]
    \centering
    \begin{subfigure}{0.35\textwidth}
        \centering
        \includegraphics[width=\textwidth]{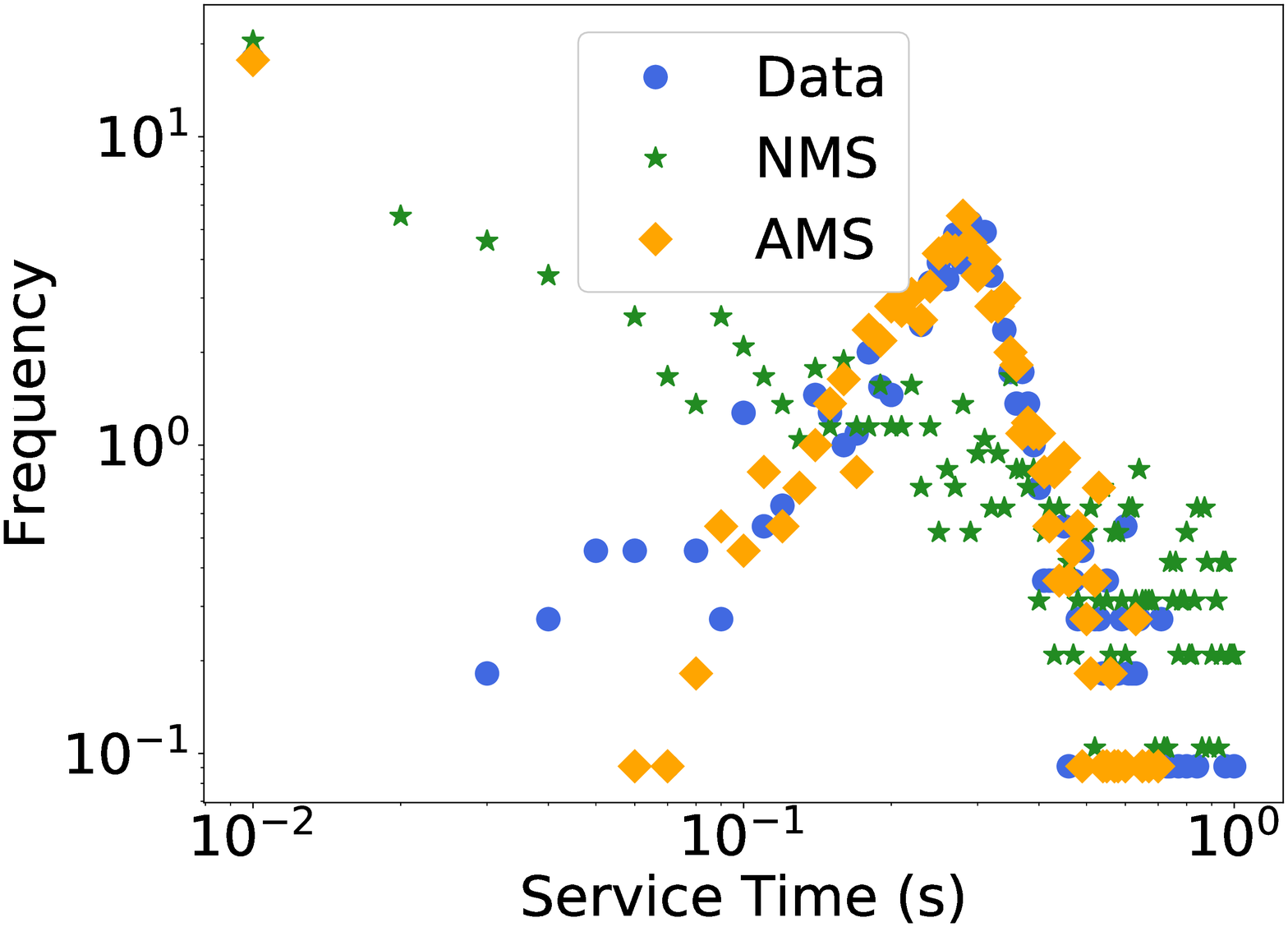}
        \caption{Mempool Service}
    \end{subfigure}%
    \begin{subfigure}{0.35\textwidth}
        \centering
        \includegraphics[width=\textwidth]{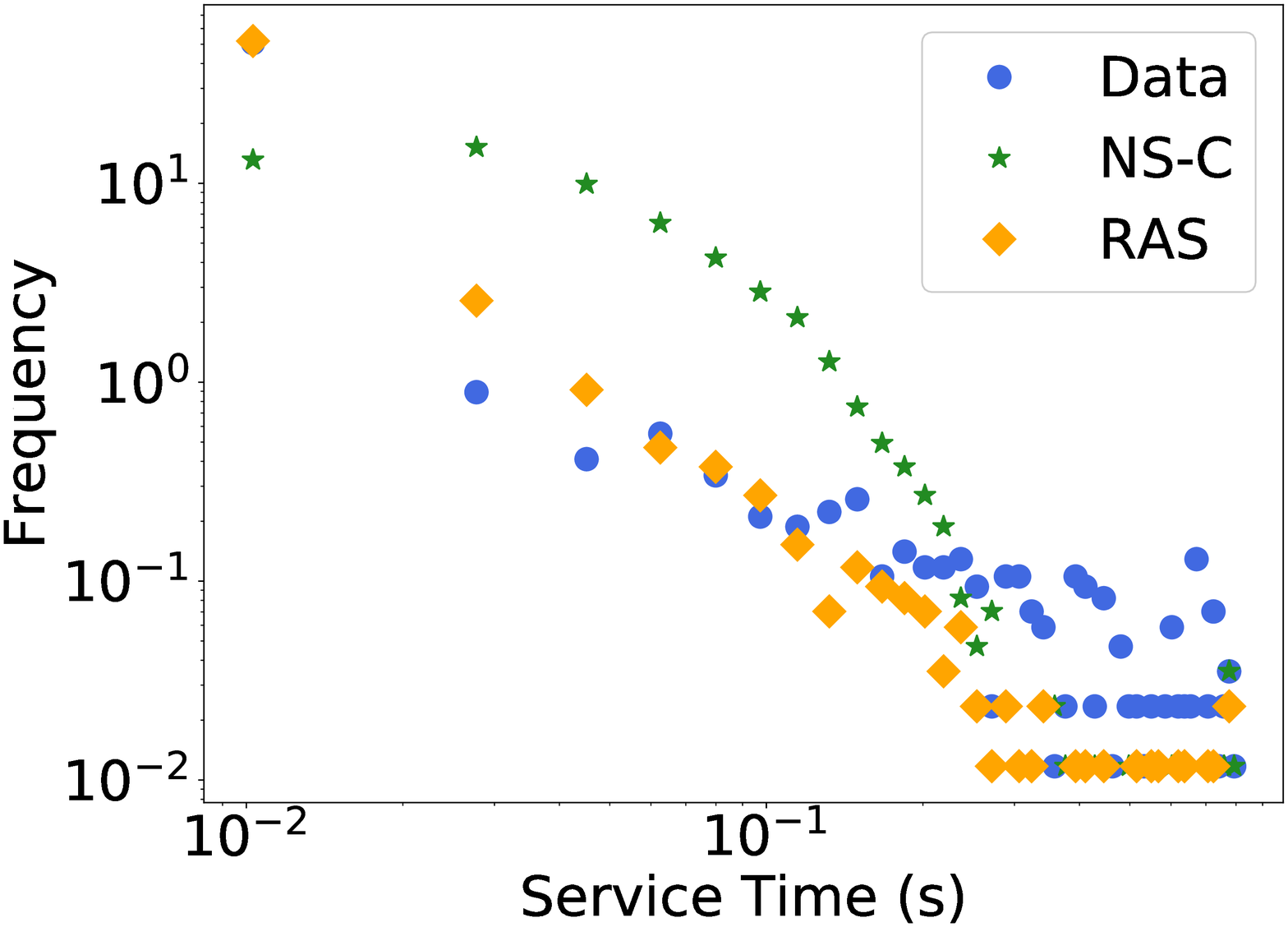}
        \caption{Github}
    \end{subfigure}
    \caption{Comparison of the probability distributions obtained from different models}
    \label{fig:model_comparision}
\end{figure*}

In order to quantify our model performance, we focus on two aspects, namely, its predictive capabilities and the ability of the model to uncover rich distributional forms in the service time. We consider the prediction error defined as $1/N \sum_i \lvert s_i - \langle \tilde{s_i} \rangle \rvert$, where $s_i$ denotes the empirical value and  $\langle \tilde{s_i} \rangle$ denotes the prediction obtained by Monte Carlo sampling. To quantify the descriptiveness of the obtained distributional form, we calculate the Kolmogorov-Smirnov (KS) statistics between the empirical and the predictive distribution and also provide Q-Q plots, against the empirical distributions in the test dataset. 

The comparison of the different models based on the predictive error and the KS statistics for the synthetic datasets are shown in Table \ref{tab:synthetic_results_table}; those for the empirical dataset are shown in Table \ref{tab:empirical_results_table}. For most datasets, the adversarial solutions (RAS, AS) outperform the neural (NS) baseline models w.r.t. KS values and prediction metric. This shows a clear advantage of these non-parametric solutions. It is important to note that the neural models strongly benefit from the flexible neural network parametrizations of the output distributions. This additional flexibility enhances the predictive power of the theoretical form and provides a fair comparison against the non-parametric adversarial solutions.

Comparing the adversarial solutions against each other, the RAS model outperforms the AS models. This indicates an improvement of the models with the inclusion of independent dynamics for the service time distribution. The RAS-NH corresponds to our RAS model where we removed the customers' arrivals hidden states dependency from the service time distribution. The difference to the RAS model highlights the importance of providing a dynamical encoding of customers' dynamics.

%This indicates an improvement of the models with the inclusion of independent dynamics for the service time distribution. Therefore the NS model is a natural extension of the theoretical ones and serves as a baseline in the predictive error Table \ref{tab:synthetic_results_table} and \ref{tab:empirical_results_table}. The RAS-NH corresponds to our RAS model where we removed the customer's arrivals hidden states dependency from the service time distribution. The difference with the RAS model highlights the importance of providing a dynamical encoding of the customer's dynamics. The service distributions improve by informing the model with the customer's behavior. The RAS model outperforms the NS and the AS models regarding the KS values in most of the datasets as well as in the prediction metric. This indicates an improvement of the models with the inclusion of independent dynamics for the service time distribution. It is important to notice that the neural models strongly benefit from the flexible neural network parametrizations of the output distributions. This additional flexibility enhances the predictive power of the theoretical form and provides a fair comparison against the non-parametric adversarial solutions.

The strength of the adversarial solution is more apparent in the distributional forms. Fig. \ref{fig:distributions-synthetic} and Fig. \ref{fig:empirical_service_distributions} show the histograms for our best model against the empirical distribution as well as Q-Q plots of both the best neural model and the best adversarial solution versus the empirical data. The adversarial model captures the short term behavior, as observed, in the upper left corner of the histogram plots. Note that the weakest results correspond to those of the non-linear Hawkes process in the synthetic datasets. These results reveal the inability of the RNPP model to encode non-linear information as mentioned in \cite{NeuralHawkes}.

In Table \ref{tab:mempool}, we present the results for the mempool model. The adversarial solution clearly outperforms all other models, both in the error and in the distribution shape as stated in the KS statistics.  We can see the qualitative behavior of the mempool models in Fig. \ref{fig:mempool_results}, where we compare the AMS model with the empirical distribution and the neural model (NMS-G). The success of the adversarial model is immediately apparent as it is the only model that captures the multi-modal nature of the mempool dataset distribution.

Finally, we show a direct comparison of the adversarial and the neural distribution obtained by the different models in Fig. \ref{fig:model_comparision}. Both for the mempool and the Github datasets the higher expressibility of the adversarial solutions is evident as the neural model only provides long tail behavior constrained by the distributions forms of the outputs.  

\section{Conclusion}

In this work, we presented a novel deep non-parametric solution for the service time distributions of queuing systems for general arrival distributions. Our solutions incorporate censoring of service times and rich estimates for complex distributions as well as independent dynamical representations for the service systems dynamics.  Our methodologies outperform neural results and  reproduce complex distributions for service times, providing richer representations which are able to recover multi-modal and  long tail distributions. We also presented a solution particularly tailored to the Bitcoin mempool queuing systems.

Future lines of work include incorporating richer representations for the arrival processes. In mobility systems, for example, the geographic information and user interactions can be encoded through hidden networks of relations. Moreover, one could make the response of the system explicit by providing new remaining service times with each new arrival.

\bibliographystyle{ACM-Reference-Format}
\bibliography{sample-base}
% \appendix

\end{document}